\documentclass[letterpaper, 10 pt, conference]{ieeeconf}  % Comment this line out if you need a4paper

\IEEEoverridecommandlockouts                              % This command is only needed if 
\usepackage{graphics} % for pdf, bitmapped graphics files
\usepackage{epsfig} % for postscript graphics files
\usepackage{times} % assumes new font selection scheme installed
\usepackage{amsmath} % assumes amsmath package installed
\usepackage{amssymb}  % assumes amsmath package installed
\usepackage{amsfonts}
\usepackage{multirow}
\usepackage{booktabs}
\usepackage{graphicx}
\usepackage{subfigure}
\usepackage{mwe}
\usepackage{threeparttable}
\usepackage[ruled,lined,linesnumbered]{algorithm2e}

\usepackage{ulem}

\usepackage{color}

\usepackage{cite}
\usepackage{hyperref}
\hypersetup{
    colorlinks=true,          % 开启链接颜色
    linkcolor=blue,           % 内部链接颜色
    citecolor=blue,           % 文献引用链接颜色
    urlcolor=blue,            % URL 链接颜色
    pdfborderstyle={/S/U/W 1} % 设置边框样式为下划线
}

\usepackage{CJKutf8}

\usepackage{multirow}

\title{\LARGE \bf
BEINGS: Bayesian Embodied Image-goal Navigation \\ with Gaussian Splatting
}

\author{Wugang Meng, Tianfu Wu, Huan Yin and Fumin Zhang% <-this % stops a space
\thanks{The work described in this paper was fully supported by grants AoE/E-601/24-N, 16203223, and C6029-23G from the Research Grants Council of the Hong Kong Special Administrative Region, China.}% <-this % stops a space
\thanks{The authors are with the Department of Electronic and Computer Engineering, Hong Kong University of Science and Technology, Hong Kong SAR. E-mail: wugang.meng@connect.ust.hk, tianfu.wu@connect.ust.hk, eehyin@ust.hk, eefumin@ust.hk}
\thanks{Corresponding author: Fumin Zhang}
}

\begin{document}

\maketitle
\thispagestyle{empty}
\pagestyle{empty}

%%%%%%%%%%%%%%%%%%%%%%%%%%%%%%%%%%%%%%%%%%%%%%%%%%%%%%%%%%%%%%%%%%%%%%%%%%%%%%%%
\begin{abstract}
Image-goal navigation enables a robot to reach the location where a target image was captured, using visual cues for guidance. However, current methods either rely heavily on data and computationally expensive learning-based approaches or lack efficiency in complex environments due to insufficient exploration strategies. To address these limitations, we propose Bayesian Embodied Image-goal Navigation Using Gaussian Splatting, a novel method that formulates ImageNav as an optimal control problem within a model predictive control framework. BEINGS leverages 3D Gaussian Splatting as a scene prior to predict future observations, enabling efficient, real-time navigation decisions grounded in the robot’s sensory experiences. By integrating Bayesian updates, our method dynamically refines the robot's strategy without requiring extensive prior experience or data. Our algorithm is validated through extensive simulations and physical experiments, showcasing its potential for embodied robot systems in visually complex scenarios. Project Page: \url{www.mwg.ink/BEINGS-web}.
% Our algorithm is validated through extensive simulations and physical experiments implemented with Graphics Processing Units (GPUs), demonstrating that RF-MPPI significantly improves navigation success rates and computational efficiency, showcasing its potential for embodied robot systems in visually complex scenarios.
\end{abstract}

%%%%%%%%%%%%%%%%%%%%%%%%%%%%%%%%%%%%%%%%%%%%%%%%%%%%%%%%%%%%%%%%%%%%%%%%%%%%%%%%
\section{Introduction} 
\label{sec:Introduction}

Visual images serve as a natural language for navigation, enabling humans to reach a target more effectively than through textual descriptions~\cite{maguire1998knowing}. In robotics, this concept is represented in Image-goal navigation (ImageNav), where visual perception guides a robot to the location where a target image was originally captured~\cite{yadav2023ovrl,wu2022image}. Robotic ImageNav encounters significant challenges in real-world and complex environments, particularly when the target is out of sight or when obstacles are present. An effective ImageNav model must enable a robot not only to navigate directly to a target but also to explore efficiently in search of it~\cite{kim2010provably}. Due to its potential applications in search and rescue operations and home service robotics, ImageNav has garnered considerable attention.

% Existing ImageNav methods fall into two categories: \textit{Learning-based approaches}, such as OVRL~\cite{yadav2023ovrl} and PoliFormer~\cite{zeng2024poliformer}, rely on reinforcement learning but face high computational costs and limited generalization to new environments. \textit{Exploration-based methods}, like Frontier~\cite{yamauchi1997frontier} and Stubborn~\cite{luo2022stubborn}, improve adaptability but lack learning mechanisms, resulting in inefficiencies in complex settings.
Existing studies for ImageNav can be broadly classified into two categories: \textit{1) Learning-based ImageNav} methods, such as OVRL~\cite{yadav2023ovrl} and PoliFormer~\cite{zeng2024poliformer}, utilize end-to-end reinforcement learning frameworks to train robots in exploration and navigation policies. These approaches typically require highly realistic simulators~\cite{szot2021habitat,straub2019replica} and millions of training trials, resulting in high computational costs and poor generalization to unseen environments. Although modular networks like NSNR~\cite{hahn2021no} and RNR-Map~\cite{kwon2023renderable} attempt to reduce simulator dependency by training on offline datasets, they continue to struggle with overfitting and generalization capabilities.
\textit{2) Exploration-based ImageNav} methods, in contrast, reduce data dependency and improve adaptability to new environments with minimal or no reliance on data-driven modules. However, these approaches have their drawbacks. Classical methods, such as Frontier~\cite{yamauchi1997frontier} and Stubborn~\cite{luo2022stubborn}, lack learning mechanisms that enable robots to refine their navigation strategies based on past experiences. Advanced methods, like MOPA~\cite{Raychaudhuri_2024_WACV} and GaussNav~\cite{lei2024gaussnav}, maintain an informative semantic map of the explored area to enhance decision-making. However, as environments become increasingly complex, these semantic maps can lead to reduced navigation efficiency and higher ImageNav costs.

%approaches~\cite{yamauchi1997frontier,bansal2020combining,luo2022stubborn,Raychaudhuri_2024_WACV}, in contrast, mitigate some of the data dependency issues faced by Learning-based ImageNav. But they are not without their own limitations. 
% design interpretable modules for pross-aware navigation and exploration without requiring training data. However, they necessitate intensive image observations for target search, resulting in lower efficiency compared to learning-based approaches.
% SLING~\cite{wasserman2023last} needs off-the-shelf learned policy that uses semantic priors to explore the scene, and the latest MOPA~\cite{Raychaudhuri_2024_WACV} introduces heuristic search algorithms for efficient exploration-exploitation but requires the use of another RL module for navigation. 
% ZSEL~\cite{al-halah2022zsel} and XGX~\cite{wasserman2024exploitation} learn policies from other visual navigation tasks.

\begin{figure}[t]
    \centering
    \includegraphics[width=0.9\linewidth]{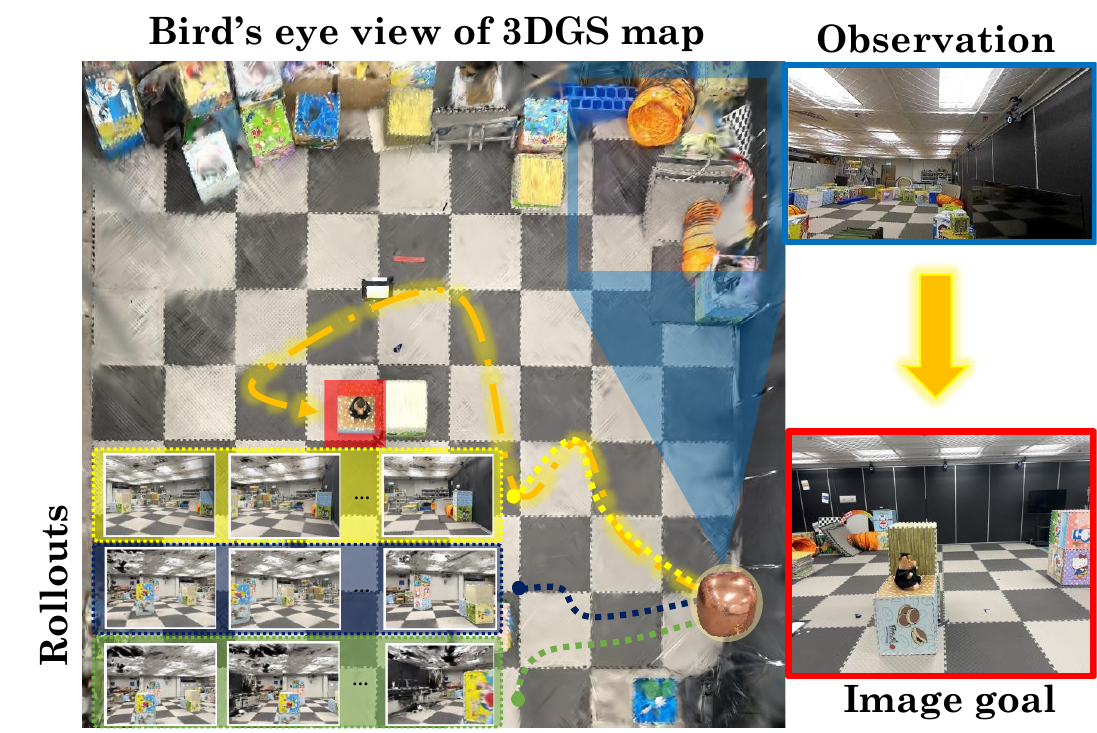}
    % \caption{Comparison of Embodied methods, Modular methods, Zero-shot methods and Experience-based methods. }
    \caption{\textbf{A schematic diagram of BEINGS.} The bird’s eye view shows a 3DGS map that the robot uses to navigate toward a target image. The robot estimates the target’s location using Bayesian principles, based on the similarity between its current observation (top right) and the image goal (bottom right). The yellow, blue, and green dotted lines show predicted rollouts, with images in corresponding color blocks showing potential future observations rendered by 3DGS (left lower). The orange dash-dotted line represents the optimal rollout selected for navigation.}
    % , and at each step, the BEINGS samples several rollouts according to the target location estimation. The 3DGS map renders observation image sequences on each rollout (left bottom corner). And the MC-BLMPC will pick out the best rollout to execute.}
    \label{fig:teaser}
\end{figure}
\begin{figure*}[t]
    \centering
    \includegraphics[width=\linewidth]{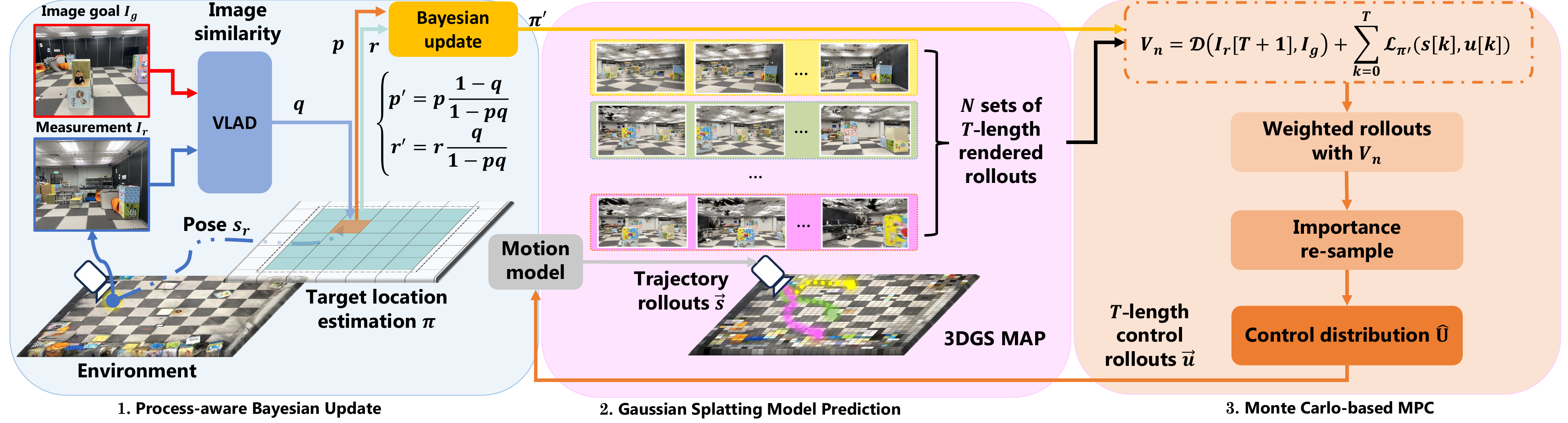}
    \caption{\textbf{System overview of BEINGS for image-goal navigation.} When the robot acquires a new image as a measurement, BEINGS updates its estimate of the target image location's distribution $\pi$ by utilizing the image similarity between the measurement and the image goal, adhering to Bayesian principles. Subsequently, it executes Monte Carlo-based MPC by sampling $N$ control sequences from the current control distribution. Using the robot's motion model and the 3DGS map, it generates $N$ image sequences of length $T$. Each image sequence is scored, and the control distribution is resampled based on these scores, incrementally approximating the optimal control distribution to guide the robot toward the target image.}
    \label{fig:system}
\end{figure*}

% \yh{Clearly point out the gaps and how to fill}
These limitations underscore a critical gap in ImageNav research: the need for a method that combines the efficiency of learning-based approaches with the robustness and adaptability of exploration-based strategies, without relying on extensive prior experience data. To address this gap, we introduce Bayesian embodied ImageNav with Gaussian splatting (BEINGS), in which the ImageNav is re-formulated as an optimal control problem. Specifically, BEINGS leverages the view synthesis function of 3D Gaussian Splatting (3DGS)~\cite{kerbl20233d} as a scene prior. Image similarities between rendered and observed images are integrated into a model predictive control (MPC) framework. The insights are twofold: first, 3DGS enables the robot to render potential future observations, informing action selection in the MPC framework to efficiently navigate to the target; second, Bayesian updates refine the robot’s strategy with each observation and past information, thus enhancing the ImageNav without external experiences. 
% Resampling control commands toward optimal outcomes further reduces computational load and improves performance in complex environments. And 
Intuitively, BEINGS is ``embodied'' because it bases navigation on real-time sensory data and the robot's interactions with its environment. It enables the robot to render future views from different positions, akin to human cognitive processes of ``imagination'' and ``memory'', and make navigation decisions dynamically informed by real-time visual perception.
% By leveraging 3DGS, the robot predicts future observations, simulating potential trajectories akin to human imagination. 
% These limitations reveal a critical gap in current ImageNav research: the need for a method that combines the efficiency of learning-based approaches with the reduced data dependency and robustness of exploration-based techniques. To address this gap, we propose a novel system, Bayesian Embodied Image-goal Navigation Using Gaussian Splatting (BEINGS), which aims to achieve efficient image-goal navigation without relying on extensive prior experience or data. Our approach reformulates ImageNav as an optimal control problem and employs a model predictive control (MPC) framework that integrates image comparisons, novel view synthesis, and process-aware information through Bayesian updates. BEINGS enables real-time adaptation and learning through experience, mimicking human-like imagination and memory. This capability allows the robot to dynamically adjust its navigation strategy based on the visual information it encounters, effectively enhancing navigation in complex environments.
Overall, our key contributions are:
\begin{itemize}
    \item Reformulating ImageNav as an optimal control problem and solving it using a Monte Carlo-based MPC.
    \item Introducing 3D Gaussian Splatting (3DGS) as a scene prior to enable efficient prediction of future observations and enhance ImageNav.
    \item Reducing data dependency and allowing dynamic adjustment of ImageNav strategies using Bayesian updates informed by real-time data.
\end{itemize}

    % \item Proposing a novel system that achieves efficient image-goal navigation without relying on extensive prior experience or data.

\section{Problem Formulation}

% four-degree-of-freedom (4DoF), encompassing three-dimensional location and roll angular orientation

Inspired by our previous study~\cite{li2021bayesian}, we formulate the ImageNav as an optimal control problem. 
Assume that the motion model $\mathcal{F}$ and the measurement model $\mathcal{H}$ are known,
we define the search space $\mathbb{S}$ as a continuous, closed, and bounded Borel set. The goal image $I_g$ is captured from the pose $s_g \in \mathbb{S}$,
% \yh{in problem formulation, regard the robot as an agent / point, without describing the pose or degree of freedom}
which indicates the source location for the goal image in the search space. We can derive the overall cost during the control sequence as a discrete Hamilton-Jacobi-Bellman (HJB) equation, stated as follows:
\begin{equation}
    J_{\pi}(s[0],U_T) =  \mathcal{D}(I_r[T+1],I_g) + \sum_{k=0}^T \mathcal{L}_{\pi}(s[k],u[k])
    \label{eq:HJB}
\end{equation}
where $\mathcal{D}(I_r,I_g)$ is the dissimilarity between the goal image $I_g$ and the image captured in terminal robot pose $I_r$, and $\mathcal{L}_{\pi}(s[k],u[k])$ signifies the one-step exploration cost~\cite{li2023integrated} incurred by implementing $u[k]$ at pose $s[k]$, given that the distributional estimation of the source location is $\pi$. To present $\pi$, the search area is divided into cells of equal volume $\{\mathbb{S}_1,\dots,\mathbb{S}_M\}$, with each cell being treated as a Borel subset of the search space $\mathbb{S}$. 
Since the measure of a specific pose is $0$ in the 4DoF continuous space, 
    we use the image similarity as a criterion for judging whether a robot navigates to a goal or not, and a threshold $\epsilon$ is established. The successful navigation is determined by the robot reaching the space of poses that generate similar images, denoted as $\mathbb{S}_\text{sim} = \{s_r \in \mathbb{S} | \mathcal{D}(I_g,I_r) < \epsilon \}$. This subset can be conceptualized as a similarity manifold surrounding the target pose, and the probability mass of $s[i]$ can be denoted as:
\begin{equation}
\begin{aligned}
    \mathcal{P}_{\pi}(s[k]) &= P_{\pi}(\mathbb{S}_\text{sim} \subset \mathbb{S}_i) ,\quad s_k \in \mathbb{S}_i\\  
    % &= P_{\pi}(s_g \subset \mathbb{S}_i)  
    &= p_i[k]
    \label{eq:prob_mass}
\end{aligned}
\end{equation}
\par In the optimal search theory~\cite{assaf1985optimal,blackwell1965discounted,kelly1982optimal,matula1964periodic}, given a movement cost function  $\{\mathcal{C}(\cdot,\cdot):\mathbb{S} \times \mathbb{S} \mapsto [0,\infty]\}$ and a priori success rate for the search target in the Borel subset $\{\mathcal{Q}(\cdot):\mathbb{S} \mapsto [0,1]\}$, the optimal search strategy at each time step is to minimize the one-step exploration cost, stated as:
\begin{equation}
    \mathcal{L}_{\pi}(s[k],u[k]) = \frac{\mathcal{C}(s[k],u[k])}{\mathcal{P}_{\pi}(s[k+1]) \mathcal{Q}(s[k+1])}
    \label{eq:box}
\end{equation}
\par Finally, the ImageNav problem can be formulated by finding the optimal control sequence $U^*_T$ that minimizes the overall cost at terminal time $T$ start from state $s[0]$:
\begin{align}
    % V(s[0],T) &= \min_{U_T} J(s[0],U_T) \label{eq:cost-to-go} \\
    U^*_T &= \arg\min_{U_T} J(s[0],U_T) \label{eq:prob_mass}
\end{align}
% This problem involves the distributional estimation of the source location $\pi$ which must be updated as new measurements are collected.
\par The distributional estimation of the target location $\pi$ must be updated as new measurements are collected. Also, the target cost equation neither exhibits a gradient nor is necessarily convex. To address these, we consider utilizing BEINGS to identify a feasible solution in practice.

% However, unlike typical signal source-seeking problems, the source location distribution for the image goal cannot be updated through a prior differentiable signal field function. We propose utilizing a Perception-Imagination-Execution (PIE) scheme to identify feasible solutions in practice. 

\section{Methodology}
% The BEINGS embodied navigation leverages process-aware Bayesian search theory, image rendering and model predictive control to navigate the blimp efficiently to the target pose estimated from target image. 
\subsection{Process-aware Bayesian Update}
The first phase of the BEINGS is utilizing new measurements to estimate the target location distribution $\pi$ of the Image-Goal. We refer to this approach as the process-aware Bayesian update~\cite{li2021bayesian}.  
With the provided definitions in Equation \eqref{eq:box}, we can update the distribution estimate $\pi$ of the source location using Bayesian search theory~\cite{richardson1971scorpion} when new measurements are obtained. At each time step $k$, if the robot in the Borel subset $\mathbb{S}_i$ observes an image but no similarity manifold is found, the revised probability mass of this subset is determined by:
\begin{equation}
    p_i[k+1] = p_i[k]\frac{1-q_i[k]}{1-p_i[k]q_i[k]}
    \label{eq:revise}
\end{equation}
And for any of other subsets, if its prior probability is $r[k]$, the posterior probability mass is calculated as:
\begin{equation}
    r[k+1] = r[k]\frac{1}{1-p_i[k]q_i[k]}
    \label{eq:update}
\end{equation}
\par In the Equation \eqref{eq:revise} and \eqref{eq:update}, $q_i$ presents the success rate of finding the target in the subset $i$. Numerous studies in the field of visual place recognition (VPR)~\cite{hausler2021patchnetvlad,keetha2023anyloc,schubert2023vprtutorial} have demonstrated that the higher the similarity between images in the feature space, the greater the probability that their corresponding camera poses are close. Thus, we use the image similarity between the target image and the image observed in subset $i$ to approximate $q_i$. And we defined $\{\mathcal{Q}(\cdot): \mathbb{S} \mapsto [0,1]\}$ as:
\begin{equation}
    \begin{aligned}
        \mathcal{Q}(s[k]) 
        % &= P_{\pi}(s[k] \subset \mathbb{S}_\text{sim} |\mathbb{S}_\text{sim} \subset \mathbb{S}_i, s[k-1] \not\subset \mathbb{S}_\text{sim} ) \\ 
        &= 1 - \mathcal{D}(I_g,\mathcal{H}(s[k]))  = q_i[k]
        \label{eq:onshot}
    \end{aligned}
\end{equation}
where $\mathcal{D}$ is the distance between the descriptors that generated by VLAD~\cite{hausler2021patchnetvlad}, which is a classical VPR method.

% Assume that a cost function $\{\mathcal{C}(\cdot,\cdot):\mathbb{S} \times \mathbb{S} \mapsto [0,\infty]\}$ is provided to introduce the the cost for the blimp to execute command $u[k]$ at state $s[k]$, and the Image-Goal is hidden in one of Borel subset stationary.
% the one-step search cost $\mathcal{L}_{\pi}$ defined in \eqref{eq:HJB} can be stated as:
% \begin{equation}
%     \mathcal{L}_{\pi}(s[k],u[k]) = \frac{\mathcal{C}(s[k],u[k])}{\mathcal{P}_{\pi}(s[k+1]) \mathcal{Q}(s[k+1])}
%     \label{eq:box}
% \end{equation}
% It has been proven~\cite{assaf1985optimal,blackwell1965discounted,kelly1982optimal,matula1964periodic} that this one-step search cost is the optimal approach to maximize the probability of detecting the target while minimizing movement costs.

At the end of each perception phase, the one-step exploration cost, which relies on the image-goal position estimate, is updated upon receiving new measurements, and this updated function remains constant until the subsequent measurement is received.

\subsection{Gaussian Splatting Model Prediction}
\begin{figure}[t]
    \centering
    \includegraphics[width=0.85\linewidth]{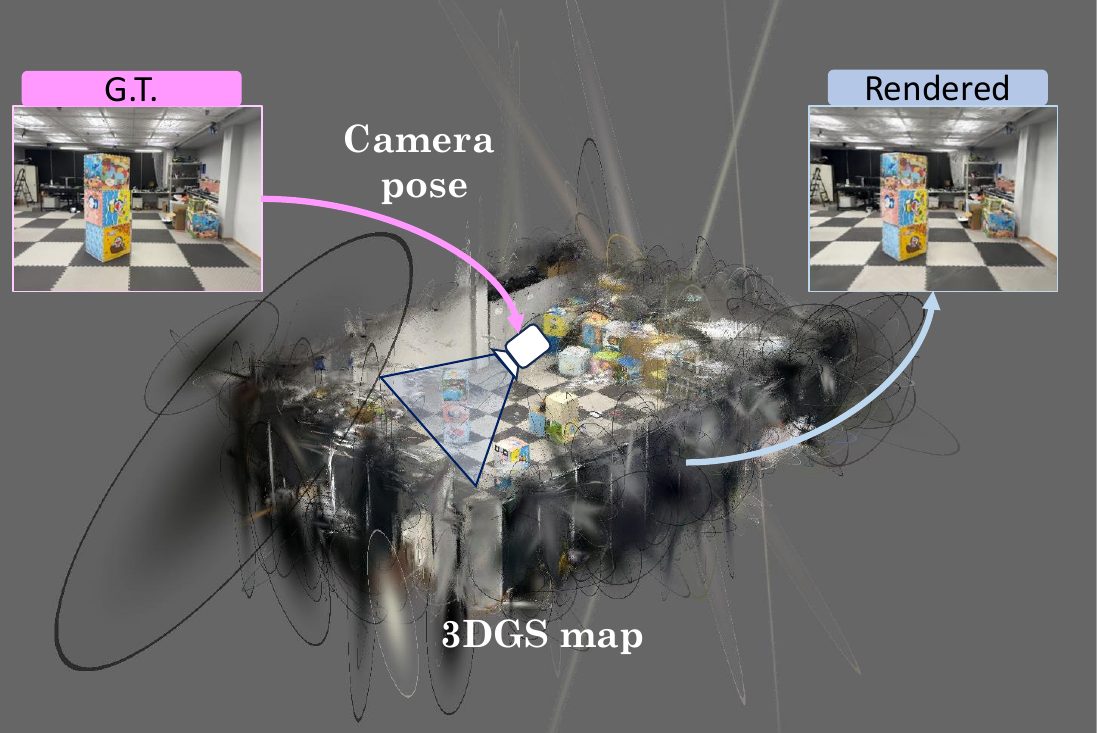}
    \caption{\textbf{Renderable radiance field map using Gaussian splatting.} Given arbitrary camera pose, 3DGS can render an image that closely resembles the real image that captured at the given pose.}
    \label{fig:render}
\end{figure}
% In the Imagination phase, through the model predictive and renderable neural radiance map, the robot can envision the paths it will take in the future and the images it will capture along those paths. 
In order to fully leverage the scene prior for exploration and navigation, we build a 3DGS as the scene prior, and a monocular camera model $\mathcal{H}$ is used to generate novel view synthesis in 3DGS. As illustrated in Figure \ref{fig:render}, the 3DGS map can dynamically predict and render the images that the camera captures in real-time for any given camera pose ~\cite{kwon2023renderable}. 
\begin{equation}
    I_r[k] = \mathcal{H}(s[k]) 
\end{equation}
then he motion function $\mathcal{F}$ as shown in Equation \eqref{eq:motion} can predict the robot's state in next time step. 
% \begin{figure}[t]
%     \centering
%     \includegraphics[width=0.8\linewidth]{figure/compare.png}
%     \caption{Examples of the measurement model and rendered image.}
%     \label{fig:render}
% \end{figure}

% With a given horizon $K$, the model predictive can forecast $N$ pairs of $K-$length trajectories $(S^1_K,\dots,S^N_K)$ and  image sequences $(I^1_K,\dots,I^N_K)$ by sampling the control rollouts $(U^1_K,\dots,U^N_K)$ from a distribution $\hat{\mathbf{U}}$. 
% The motion function of the robot $\mathcal{F}$ is characterized by nonlinearity and significant coupling between translational and rotational motions~\cite{identyfy}. However, our earlier study~\cite{tao2021swing,10587378} simplified the blimp's discrete motion model $\mathcal{F}$ as shown in \eqref{eq:motion} by incorporating a constraint on the control inputs: 
\begin{equation}
    \begin{aligned}
            s^n[k+1] &= \mathcal{F}(s^n[k],u^b[k]) %\\ 
            % \begin{pmatrix}
            %     x[k+1] \\
            %     y[k+1] \\
            %     z[k+1] \\
            %     \theta[k+1]
            % \end{pmatrix}
            % =\begin{pmatrix}
            %     x[k] \\
            %     y[k] \\
            %     z[k] \\
            %     \theta[k]
            % \end{pmatrix} &+ \begin{pmatrix}
            %     \mathbf{R}(\theta[k]) & \mathbf{0} \\
            %     \mathbf{0} & \mathbf{1} \\
            % \end{pmatrix} \begin{pmatrix}
            %     \nu_x[k] \\
            %     \nu_y[k] \\
            %     \nu_z[k] \\
            %     \omega[k] \\
            % \end{pmatrix} \\ 
            % & \mathbf{R}(\theta[k]) =  \begin{pmatrix}
            %     \cos{\theta[k]} & \sin{\theta[k]}  \\
            %     \sin{\theta[k]} & -\cos{\theta[k]}  \\
            % \end{pmatrix} 
            \label{eq:motion}
    \end{aligned} 
\end{equation}
% In this context, $\mathbf{0}$ and $\mathbf{1}$ represent the second-order zero matrix and the identity matrix, respectively. The matrix $\mathbf{R}(\theta[k])$ denotes the state-dependent transformation matrix from the blimp's body frame $b$ to the inertial frame $n$. The state $s^n=(x^n,y^n,z^n,\theta^n)$ of the blimp, illustrated in the Figure \ref{fig:frame}, includes the coordinates in the inertial frame and the yaw angle. Meanwhile, the control input $u^b=(\nu^b_x,\nu^b_y,\nu^b_z,\omega^b)$ consists of the linear velocities along the $X, Y$ and $Z$ axes in the body frame, as well as the angular velocity around the $Z$ axis. The input constraint applied states that at any specific moment, only one variable among $\nu^b_x$, $\nu^b_y$, $\nu^b_z$, and $\omega^b$ can be non-zero.  
% \begin{figure}[t]
%     \centering
%     \includegraphics[width=0.5\linewidth]{figure/frame.png}
%     \caption{The control coordinates and the state coordinates of the blimp.}
%     \label{fig:frame}
% \end{figure}

With a given horizon $K$, the model predictive can forecast $N$ pairs of $K-$length trajectories $(S^1_{1:K},\dots,S^N_{1:K})$ and image sequences $(I^1_{1:K},\dots,I^N_{1:K})$ by sampling the control rollouts $(U^1_{1:K},\dots,U^N_{1:K})$ from a distribution $\hat{\mathbf{U}}$. So, at a specific time step $k$, we can calculate the overall cost (in Equation \eqref{eq:HJB}) of the rollout $i$ with the corresponding predictive trajectory $S^i_{k+1:K+k}$ and image sequence $I^i_{k+1:K+k}$ by:
\begin{equation}
    \mathcal{J}_{\pi}(s[k],k+K) = \mathcal{D}(I^i_{K+k},I_g) + \sum^{k+K-1}_{t=k}\mathcal{L}_{\pi}(s^i[t],u^i[t])
\end{equation}
in which $s^i[k] = s[k]$. For each rollout, we define its unnormalized weight by $w^k_i = e^{-\mathcal{J}_{\pi}(s[k],k+K)}$.
% \begin{equation}
%     w^k_i = e^{-\mathcal{J}_{\pi}(s[k],k+K)}
%     \label{eq:weight}
% \end{equation}

% With the model predictive rendering, prior to implementing control commands in real world, the costs and potential rewards of executing those commands could be predicted and pre-calculated in this phase. 
% by applying Girsanov’s theorem:
% \begin{equation}
%     \begin{aligned}
%         \mathbf{U} &= \sum^N_{i=1} U^i_K w^i \\
%         w^i &= \exp{[-J(s[0],U^i_K)]}
%     \end{aligned} \label{eq:distribution}
% \end{equation}

% Prior to implementing control commands in real world, the costs and potential rewards of executing those commands are precalculated in this phase.
% \begin{equation}
%     \begin{pmatrix}
%     x[k+1] \\
%     y[k+1] \\
%     z[k+1] \\
%     \theta[k+1]
%     \end{pmatrix} = %\begin{bmatrix} x[k] \\
%     % y[k] \\
%     % z[k] \\
%     % \theta[k] \end{bmatrix}
% \begin{pmatrix}
%                 \cos{\theta[k]} & \sin{\theta[k]} & 0 & 0 \\
%                 \sin{\theta[k]} & -\cos{\theta[k]} & 0 & 0 \\
%                 0 & 0 & 1 & 0 \\
%                 0 & 0 & 0 & 1 
%             \end{pmatrix}
% \end{equation}
% The Radiance Field~\cite{mildenhall2021nerf,kerbl20233d} serves as an innovative map representation, aiding robots in pose estimation through monocular visual data~\cite{adamkiewicz2022vision,maggio2023loc,meng2024nurf} and optimizing trajectories for collision avoidance using rendered image sequences~\cite{chen2024catnips,chen2024splat}.

\subsection{Monte Carlo-based MPC}
\begin{figure}[t]
    \centering
    \includegraphics[width=0.85\linewidth]{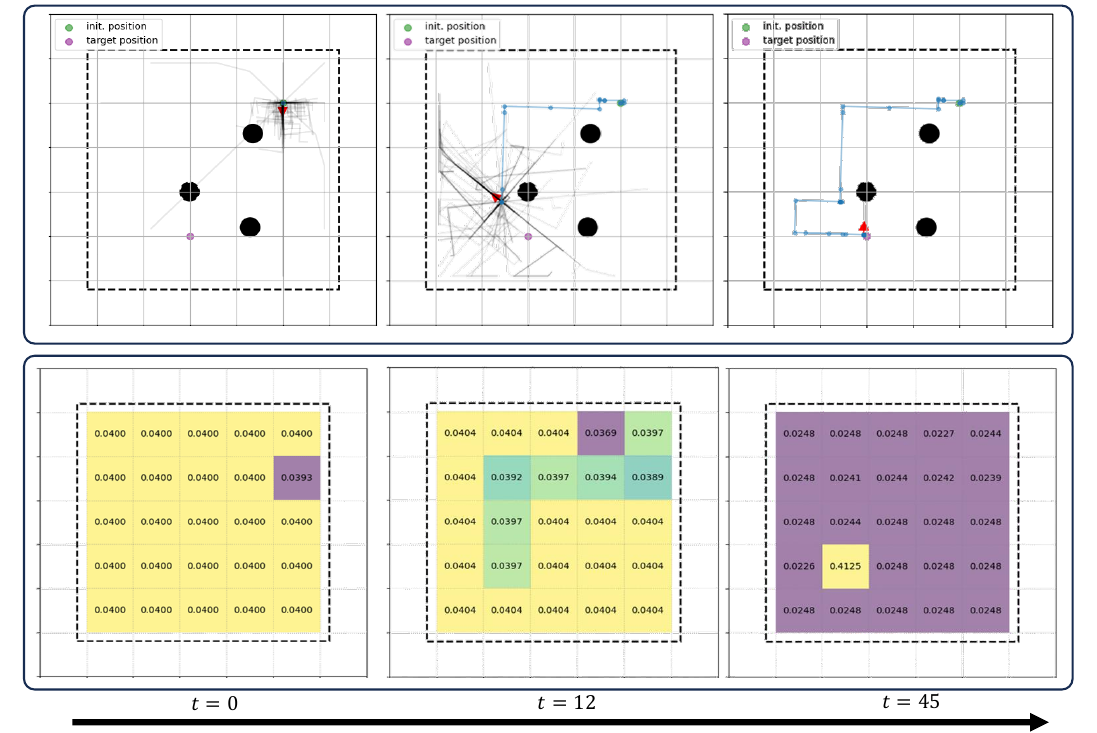}
    \caption{\textbf{ImageNav process using BEINGS.} The process shows the Monte Carlo-base MPC process (Top) and the probability of the target allocated in each $\mathbb{S}_i$ is changing with process-aware Bayesian update (Bottom). }
    \label{fig:converge}
\end{figure}
We employ a Monte Carlo method to determine the optimal control policy that minimizes the value of HJB by approximating the control distribution $\hat{\mathbf{U}}$ from which we sample control rollouts to match the optimal distribution represented by $\mathbf{U}^*$.
% using the information obtained in the Perception phase. Our goal is to approximate the control distribution $\hat{\mathbf{U}}$ from which we sample control rollouts to match the optimal distribution represented by $\mathbf{U}^*$.
Therefore, we first redefine the optimal objective in its stochastic form:
\begin{equation}
    U^*_T = \arg\min_{U_T} E_{\hat{\mathbf{U}}}[J(s[0],U_T)]
    \label{eq:exp}
\end{equation}
where $E_{\hat{\mathbf{U}}}[\cdot]$ means the expectation HJB cost of the distribution $\hat{\mathbf{U}}$. And, instead of directly solving this stochastic Equation \eqref{eq:exp}, we address the minimization problem by adjusting the probability distribution of the controls $\hat{\mathbf{U}}$ towards the optimal probability distribution $\mathbf{U}_T^*$ and use the control sequence with the lowest cost-to-go as the numerical approximation of $U^*_T$~\cite{chaslot2008cross}:
\begin{equation}
    \begin{aligned}
        \mathbf{U}^* &= \arg\min_{\hat{\mathbf{U}}} D_{KL}(\mathbf{U}^*_T \parallel \hat{\mathbf{U}})  \\
        U^*_T &= \min{U} \quad\quad\quad U \sim \mathbf{U}^*
    \end{aligned} \label{eq:cem}
\end{equation}
in which $D_{KL}(\cdot||\cdot)$ stands for the Kullback–Leibler divergence (KL divergence) between two distributions. Although, the HJB equation is updated with the robot perceptions, we note that the entire image-goal navigation follows Bellman optimality~\cite{sniedovich1986new}, ensuring satisfies with:
\begin{equation}
    \mathcal{V}(s[0],T) = \mathcal{V}(s[0],k) + \mathcal{V}(s[k+1],T) \quad \forall k \in (0,T)
    \label{eq:bellman}
\end{equation}
\par Therefore, at each time step $k$, our focus is on minimizing the global cost by reducing the KL divergence between $\hat{\mathbf{U}}_{k:k+K}$ and $\mathbf{U}^*_{k:k+K}$ through sequential importance resampling~\cite{doucet2009tutorial,Grady2016mppi}. 
% In practice, we define the unnormalized weight of each rollout as $w^k_i = e^{-\mathcal{J}(s[k],k+K)}$.
% \begin{equation}
%     w^k_i = e^{-\mathcal{J}(s[k],k+K)}
%     \label{eq:unw}
% \end{equation}
Assume we draw $N$ independent samples from $\hat{\mathbf{U}}_{k-1:k+K-1}$ then we obtain the Monte Carlo approximation of $\hat{\mathbf{U}}_{k:k+K}$ as:
\begin{equation}
    \hat{\mathbf{U}}_{k:k+K} = \sum^N_{i=1} W_i \delta_{U^i_{k:k+K}}
    \label{eq:proposed}
\end{equation}
where $\delta_{U^i_{k:k+K}}$ denotes the Dirac delta mass located at rollout $U^i_{k:k+K}$ and normalised weight of the rollouts are:
\begin{equation}
    W^k_i = \frac{w^k_i}{\sum^N_{i=1}w^k_i}
    \label{eq:nw}
\end{equation}
\par To obtain approximate samples from $\mathbf{U}^*_{k:k+K}$, we simply samples from the Monte Carlo approximation $\hat{\mathbf{U}}_{k:k+K}$; specifically, we select $\hat{\mathbf{U}}_{k+1:k+K+1}$ with probability $W^k_i$. For each time step, based on Equation \eqref{eq:bellman}, the optimal policy $u$ is the first control input of $U^*_T$. 
% \begin{equation}
%     u = U^*_T[0]
% \end{equation}
% However, this approach proves to be highly inefficient when the target distribution is updated through perception. We note that the entire image-goal navigation follows Bellman optimality~\cite{sniedovich1986new}, ensuring \eqref{eq:cost-to-go} satisfies with:
% \begin{equation}
%     \mathcal{V}(s[0],T) = \mathcal{V}(s[0],k) + \mathcal{V}(s[k+1],T) \quad \forall k \in (0,T)
%     \label{eq:bellman}
% \end{equation}
% By substituting Equation \eqref{eq:bellman} and \eqref{eq:HJB} back into Equation \eqref{eq:cost-to-go}, we can obtain the recursive formula for the minimum cost-to-go based on the HJB equation as follows~\cite{brunton2022data}:
% \begin{equation}
%     \mathcal{V}(s[k],k) = \min_u[ \mathcal{L}_{\pi}(s[k],u) + \mathcal{V}(s[k-1],k-1)]
%     \label{eq:dp}
% \end{equation}
% Thus, at any time step $k$, the the optimal control input is derived as:
% \begin{equation}
%     u = \arg\min_u[\mathcal{L}_{\pi}(s[k],u) + \mathcal{V}(s[k-1],k-1)]
%     \label{eq:policy}
% \end{equation}
By implementing the control input $u$ at the current time step and resampling after receiving new observations, our approach can quickly approximate the target distribution. 

In practice, we introduce random noise to prevent trajectory degeneracy. The complete approach of BEINGS is briefly outlined in Algorithm \ref{alg:PIE}. A demonstration of BEINGS is shown in Figure \ref{fig:converge}. In the top image, the black solid lines represent rollouts sampled based on the distribution $\hat{\mathbf{U}}$. In the bottom image, the number at the center of each cell $\mathbb{S}_i$ indicates the probability mass of the target appearing in that cell, and all probability masses collectively form the estimate $\pi$ of the target location. It can be observed that as the robot searches the environment and gradually approaches the target location, both $\hat{\mathbf{U}}$ and $\pi$ progressively converge.

\begin{algorithm}[t]
    \SetAlgoLined
    \SetKwInOut{Input}{Input}
    \SetKwFunction{cam}{camera}
    \SetKwFunction{up}{update}
    \SetKwFunction{F}{$\mathcal{F}$}
    \SetKwFunction{IS}{Importance resample}
    \SetKwFunction{J}{$\mathcal{J_{\pi}}$}
    \SetKwFunction{elite}{$\arg\min$}
    \Input{$s[0],I_g$}
    Initialize the target distribution $\pi$;

    Initialize the control distribution $\mathbf{U}$;
    
    $s \gets s[0]$;
    
    \While{task not completed}{
        % \tcp{Perception phase}
        
        $I \gets$ \cam($s$);

        $\pi \gets$ \up($I,s$) \tcp*{Equation \eqref{eq:revise},\eqref{eq:update}}

        % \tcp{Imagination phase}
        
        $W \gets \mathbf{0}_{1:N}$;
        
        \For{Monte Carlo rollouts $n=1,\dots,N$}{
            $s'[0] \gets s$;
            
            \For{MPC horizon $k=0,\dots,K$}{
                sample $u[k]$ from $\mathbf{U}$;

                $s'[k+1] \gets$ \F($s[k],u[k]$);

                $V' \gets$ \J($s[k],k+1$);

                $W[n] \gets W[n] + \exp{(-V')} $;
            }
        }
        Normalize $W$ \tcp*{Equation \eqref{eq:nw}}

        % \tcp{Execution phase}

        $\mathbf{U} \gets$ \IS($\mathbf{U},W$);
        
        Gets the best sequence $U$; 

        Apply the first input $U[0]$;
        
    }
    \caption{BEINGS}
    \label{alg:PIE}
\end{algorithm}

\section{Experiments}
\begin{table*}[t]
    \centering
    \caption{Experimental Results with Different Evaluation Metrics and Tasks}
    \renewcommand\arraystretch{1.2}
    \resizebox{\linewidth}{!}{
\begin{tabular}{lcccccccccccccc}
\hline
\hline
 & \multirow{2}{*}{Scene Priors}      & \multirow{2}{*}{Exploration Strategies} & \multicolumn{4}{c}{Easy}        & \multicolumn{4}{c}{Medium}     & \multicolumn{4}{c}{Hard}        \\ \cline{4-15} 
 &                                    &                                     & SR(\%)$\uparrow$ & NS$\downarrow$ & SPC$\uparrow$  & NE(m)$\downarrow$ & SR(\%)$\uparrow$ & NS$\downarrow$ & SPC$\uparrow$  & NE(m)$\downarrow$ & SR(\%)$\uparrow$ & NS$\downarrow$  & SPC$\uparrow$  & NE(m)$\downarrow$ \\ \hline
 & \multirow{2}{*}{VPR Database~\cite{keetha2023anyloc}}       & Directly                         &       11 &    \textbf{1}    &   0.71    &  2.16 &       68 &      4 &      0.07&      2.88 &    64&       6  &      0.00 &    2.61\\
 &                                    & FB~\cite{yamauchi1997frontier}                                  &    \textbf{100} &       5 &      0.87 &       2.16 &     \textbf{100} &       37 &   0.56&    2.88&    $-$&     $-$&  $-$&    $-$\\ \cline{3-15} 
 & \multirow{4}{*}{Semantic Grid Map~\cite{Raychaudhuri_2024_WACV}} & Directly  &       2 &      \textbf{1} &      0.00 &    \textbf{0.20}   &      2&       5 &    0.03  &      \textbf{0.20} &   2     &     7 &   0.00   &   \textbf{0.20}    \\
 &                                    & FB~\cite{yamauchi1997frontier}                                  &       34 &     22&      0.36 &     0.36  &     46&      40&      0.34&      \textbf{0.20} &    $-$     &      $-$    &   $-$    &     $-$  \\
 &                                    & Stubborn~\cite{luo2022stubborn}                            &        25&        22&      0.21&     0.28&    34&        35&      0.28&      0.24 &     $-$    &      $-$    &    $-$   &    $-$    \\
 &                                    & Uniformly~\cite{Raychaudhuri_2024_WACV}                           &        26&        18&      0.22&    0.22&        56&      42 &    0.17&      0.24 &      $-$  &      $-$   &    $-$  &    $-$   \\ \cline{3-15} 
 & \multirow{2}{*}{3DGS~\cite{xu2024splatfacto}}              & Directly                        &  38  &       \textbf{1} &       0.28&      2.55&        6&        \textbf{2} &      0.03&       2.75&        5.11&        \textbf{5} &      0.00 & 4.25       \\
 &                                     & BEINGS(ours)                      & \textbf{100}    & 2 & \textbf{0.98} & 1.91      & \textbf{100}     & 7 & 0.73 &  2.52     & \textbf{78}     & 12 & \textbf{0.53} &  2.55  \\ 
 \cline{2-15} 
 \hline
 \hline
\end{tabular}}
\label{tab:simulation}
\end{table*}

% We report results for image-goal navigation both in simulation and real-world scenes.
\subsection{Experiment Setup}
We conduct experiments on image-goal navigation in real-world scenes, verifying the proposed BEINGS on a miniature blimp robot flying in an indoor test field. The test field measures 10 meters in length, 10 meters in width, and 2 meters in height, offering ample space for navigation. 
\begin{figure}[t]
    \centering
    \includegraphics[width=0.85\linewidth]{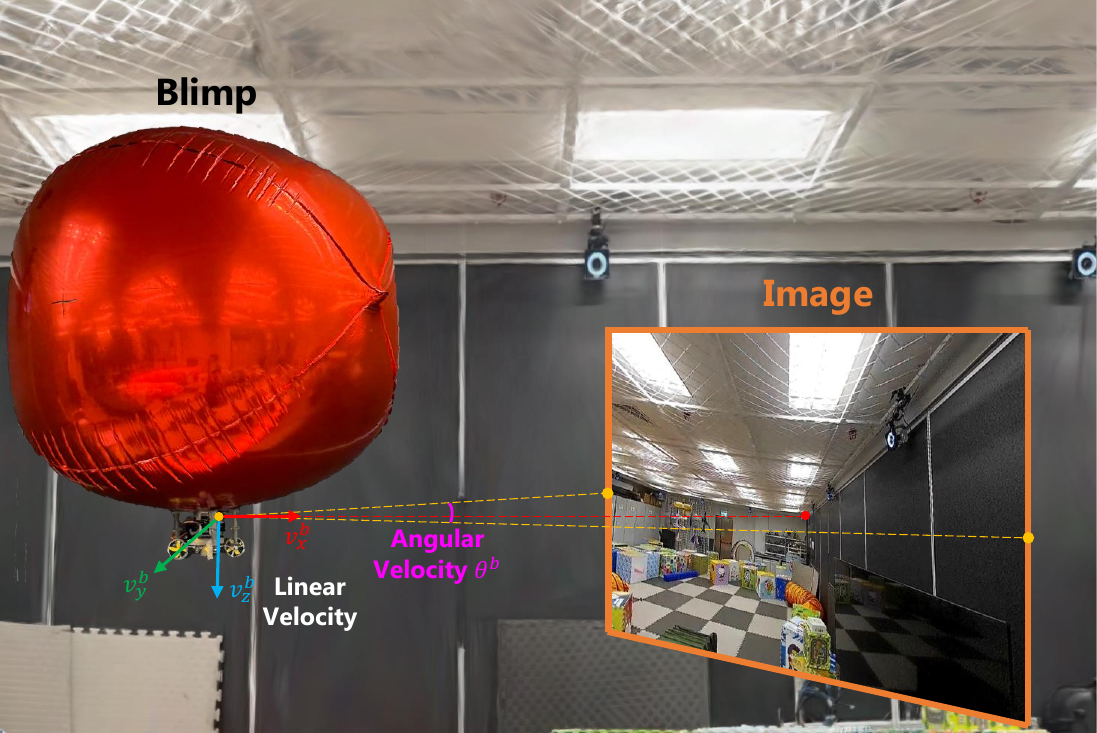}
    \caption{\textbf{Miniature blimp robot.}  In this study, the body frame is set as the camera frame, and control commands are applied to the body frame for navigation.}
    \label{fig:frame}
\end{figure}

\subsubsection{Robot Platform}

The blimp robot is equipped with a monocular camera for real-time image capture and a digital image transmitter that sends image stream to computer for processing. Figure~\ref{fig:frame} showcases our blimp robot operating in the indoor test field. Theoretically, the motion $\mathcal{F}$ is characterized by nonlinearity and is coupling between translational and rotational motions~\cite{tao2020modeling}. Our recent studies~\cite{tao2021swing,meng2024hybrid} simplify the discrete motion function as:
\begin{equation}
    \begin{pmatrix}
                x[k+1] \\
                y[k+1] \\
                z[k+1] \\
                \theta[k+1]
            \end{pmatrix}
            =\begin{pmatrix}
                x[k] \\
                y[k] \\
                z[k] \\
                \theta[k]
            \end{pmatrix} + \begin{pmatrix}
                \mathbf{R}(\theta[k]) & \mathbf{0} \\
                \mathbf{0} & \mathbf{1} \\
            \end{pmatrix} \begin{pmatrix}
                \nu_x[k] \\
                \nu_y[k] \\
                \nu_z[k] \\
                \omega[k] \\
            \end{pmatrix} \\ 
            % & \mathbf{R}(\theta[k]) =  \begin{pmatrix}
            %     \cos{\theta[k]} & \sin{\theta[k]}  \\
            %     \sin{\theta[k]} & -\cos{\theta[k]}  \\
            % \end{pmatrix} 
\end{equation}
where $\mathbf{0}$ and $\mathbf{1}$ represent the second-order zero matrix and the identity matrix, respectively. The matrix $\mathbf{R}(\theta[k])$ represents the state-dependent transformation from the body frame $b$ to the inertial frame $n$. The state $s^n=(x^n,y^n,z^n,\theta^n)$ of the blimp includes the coordinates in the inertial frame and the yaw angle. The control input $u^b=(\nu^b_x,\nu^b_y,\nu^b_z,\omega^b)$ consists of linear velocities along the $X, Y,$ and $Z$ axes in the body frame, as well as the angular velocity around the $Z$ axis. The input constraint specifies that at any given moment, only one variable among $\nu^b_x$, $\nu^b_y$, $\nu^b_z$, and $\omega^b$ can be non-zero.
\begin{figure*}[t]
    \centering
    \includegraphics[width=\linewidth]{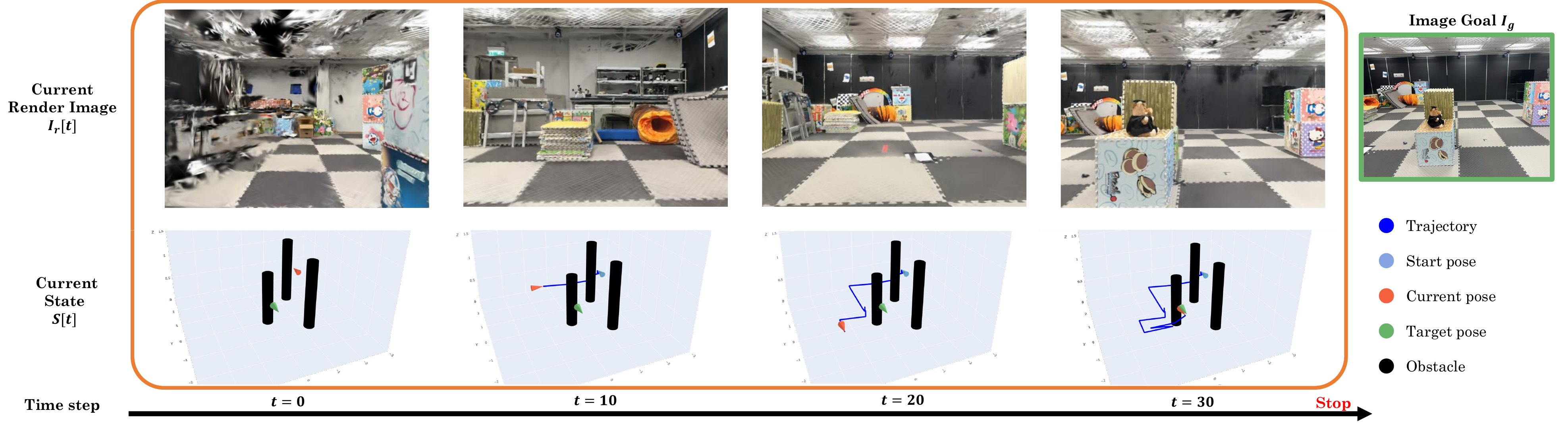}
    \caption{\textbf{One hard case study for image-goal Navigation.} The robot states and rendered images are displayed from left to right.}
    \label{fig:example}
\end{figure*}

\subsubsection{Metrics}

To evaluate performance quantitatively, we define success of ImageNav as: the task is successful if the robot enters the Last-Mile stage~\cite{wasserman2023last} within 50 steps, meaning the target is in view. The normal movement cost for the blimp is defined as 50 times the distance moved, while a collision incurs a cost of 1000. We assess different ImageNav methods using success rate, efficiency, and cost. Specifically, we report results with Success Rate (\textbf{SR}), Success-weighted Path Cost (\textbf{SPC}), absolute average Navigation Error (\textbf{NE}), and minimum Number of Steps (\textbf{NS}). Note that SPC is a strict metric derived from SPL (Success-weighted Path Length)~\cite{anderson2018evaluation}. Even if collisions occur, the robot is considered to have successfully completed navigation, but with a penalty to the total cost.
% Therefore, when evaluation is conducted in previously unexplored and fairly complex environments with obstacles, an SPC of 0.5 is considered a good level of navigation performance. Additionally, since we only consider the robot's success rate within the first 50 steps, the average number of steps is truncated at 50, and we report the minimum number of steps taken. When the SPC is greater than 0.5, the fewer the minimum number of steps used for navigation, the more we regard the algorithm as superior.

\subsubsection{Tasks by Difficulty}

We validate the BEINGS at three difficulty levels. \textbf{Easy.} No obstacles exist between the robot and the target image, requiring only orientation adjustments to locate the target. \textbf{Medium.} One 1.5m high, 0.25${}^2$ obstacle is placed between the robot and the target image. The robot can either bypass the obstacle or adjust its altitude to see the target. \textbf{Hard.} In this long-distance scenario, the robot is required to navigate past three 1.5m high, 0.25${}^2$obstacles and then adjust its orientation to search the target image. All navigation strategies are tested 50 times from a given initial pose in each task. Repetitive tests are conducted in the 3DGS map on a PC server, and real-world verification is also performed through field tests.

\subsection{Comparisons}
We compare BEINGS with advanced exploration-based methods to demonstrate its superiority in unlearned environment, considering the combinations of different scene priors and exploration strategies.
\subsubsection{Scene priors}
% In addition to 3DGS, the other two scene priors we selected are the 2D top-down semantic grid map from MOPA~\cite{Raychaudhuri_2024_WACV} and the VPR  database from AnyLoc~\cite{keetha2023anyloc}.
\textbf{VPR Database.} 
We collect 534 RGB images at a resolution of $1920 \times 1440$, along with the position and pose information of the phone during capture, using an iPhone with markers attached in an OptiTrack-equipped room. These 534 images are then converted into a 1024-dimensional vector database using the AnyLoc-VLAD-DINO~\cite{keetha2023anyloc} method. For any image goal, the VPR Database can provide the pose of the most similar image retrieved from the database~\cite{johnson2019billion}. \textbf{Semantic Grid Map.} As proposed in MOPA~\cite{Raychaudhuri_2024_WACV}, the map is a 2D top-down grid map, where each cell is a square spanning $0.2^2$ and contains the semantic label of the objects present at that location. The semantic labels for each grid cell are manually annotated. \textbf{3DGS.} Generated from the 534 images and camera poses collected in the VPR Database. 

\subsubsection{Exploration Strategies}
\textbf{Directly Approach.} Most straightforward exploration strategy, it estimates the most likely waypoint and navigates directly to it. \textbf{Frontier-based (FB).}~\cite{yamauchi1997frontier} A traditional heuristic exploration method that directs the robot to the nearest unexplored point with the highest probability. \textbf{Stubborn.}~\cite{luo2022stubborn} An exploration strategy based on fixed rules, guiding the robot to explore the four corners of a square area sequentially, expanding that area if the target is not found. \textbf{Uniformly.}~\cite{Raychaudhuri_2024_WACV} A composite exploration strategy where the robot samples an exploration goal uniformly on a top-down 2D map. A new exploration goal is resampled if the robot does not reach the target within a specific time step.
% In current state-of-the-art ImageNav algorithms, methods that do not use samples for training before the last-mile typically adopt frontier-based exploration (FBE)~\cite{krantz2023navigating} strategies. Therefore, we use three different environmental information representations: VPR database~\cite{keetha2023anyloc}, Neural Networks(NN)~\cite{yadav2023ovrl}, and 3DGS~\cite{kwon2023renderable} paired with FBE as baselines. To ensure fairness, the amount of data used to train the environmental information is consistent across all methods. Additionally, since the two map representations, apart from 3DGS, cannot provide information about obstacle locations, we use an occupancy grid map to assist FBE in obstacle avoidance. 
% name:            /      DDP       /      Frontier    / Prticle based/  PIE(ours) /
% Map:           / VPR Database   /  VPR Database   /   VPR Database / 3DGS map    /
% Size:        /     472Mb      /     472Mb      /  ------        /252Mb        / 
\subsection{Experimental Results}

The quantitative results are presented in Table \ref{tab:simulation}. Semantic grid map-based methods achieve high navigation accuracy (NE $<$ 1m) due to the fine-grained grid map. However, there is no such thing as a free lunch. The cost of excessively dense grid maps is the high NS metric, i.e., a low exploration efficiency. When using heuristic exploration strategies on this grid map, the ImageNav task cannot be completed within 50 steps in hard tasks~\cite{yamauchi1997frontier,luo2022stubborn,Raychaudhuri_2024_WACV}. Regarding navigation efficiency, the direct approaches require the fewest steps to navigate to the target point but with a low navigation success rate\cite{bansal2020combining}. In environments with obstacles, the collision penalties result in a very low SPC, approximately equal to 0 when rounded to two decimal places. Overall, our proposed BEINGS is demonstrated to complete tasks within 50 steps with a high probability across all difficulty levels while maintaining an SPC above 0.5, especially on hard tasks where other methods' performances are unsatisfactory. The results demonstrate that BEINGS can navigate to the target with a high success rate SR ($ \geq 70\%$) in a limited number of steps NS ($\leq 50$), while ensuring that the movement cost remains within an acceptable range (SPC $> 0.5$).

% if collision costs are not considered

Figure \ref{fig:example} presents a case study of long-distance ImageNav using BEINGS. We replace the real visual measurements, $I_r$, with the rendered images from the current robot pose in the 3DGS map. The BEINGS method successfully guides the robot to explore the environment, navigate around obstacles, and reach the target finally. 

\subsection{Ablation Study}

\begin{figure}[t]
    \centering
    \includegraphics[width=0.95\linewidth]{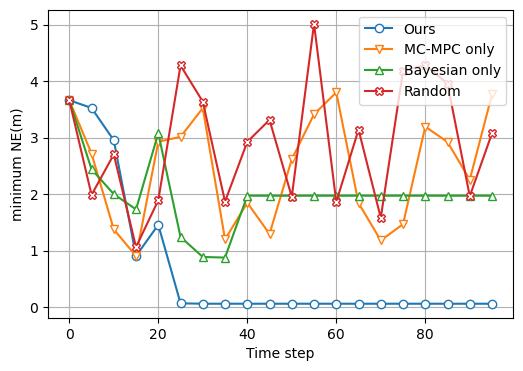} 
    \caption{\textbf{Ablation study.} The minimum value of navigation error NE in the hard scenarios for each method. Our method of using both Bayesian updating and MC-MPC converges to the nearest pose to the target in the shortest number of steps.}
    \label{fig:ablation}
\end{figure}

To validate the modules, we evaluate performance with either the Bayesian estimator or the Monte Carlo step disabled on challenging tasks. As shown in Figure \ref{fig:ablation}, performance declines whenever one of these modules is turned off, with the minimum navigation error achieved when both modules function together. When the Bayesian estimator is disabled, the MC-MPC relies solely on image similarity for resampling, causing the robot to struggle to converge to a specific location. Although using only the Bayesian estimator allows for convergence without resampling, it leads to convergence at an incorrect location, resulting in a higher navigation error. The Random approach illustrated in Figure \ref{fig:ablation} represents the scenario where both the Bayesian estimator and MC-MPC are disabled, executing rollouts randomly.

\subsection{Real-world Demonstrations}

\begin{figure}[t]
    \centering
    \includegraphics[width=0.95\linewidth]{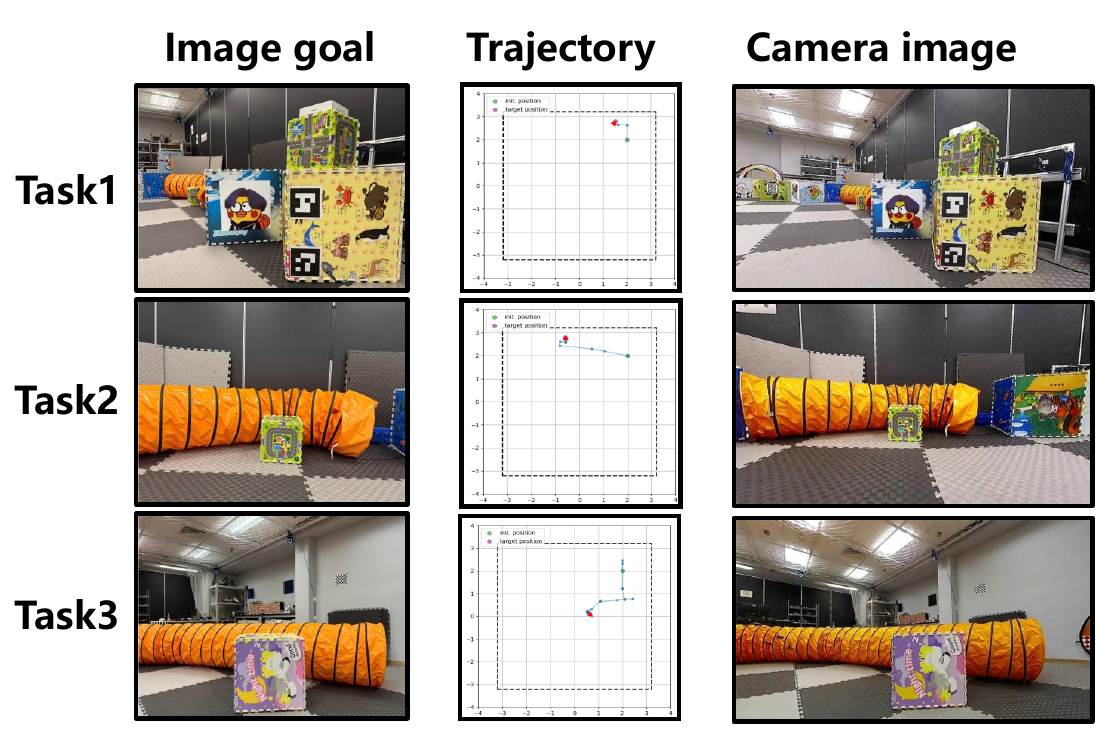} 
    \caption{\textbf{Real-world case studies.} Blimp explores the environment driven by the BEINGS algorithm (Middle) from the initial pose to the pose captured by the target image (Left) and true images viewed by the camera when algorithm converged (Right). }
    \label{fig:real}
\end{figure}

To further verify our proposed system, we conduct a real-world experiment on our blimp robot and the OptiTrack system is utilized for robot pose tracking. We pre-traine a 3DGS model, leveraging metric-scaled poses and RGB images from a handheld camera. The experimental setup consisted of a blimp equipped with a $1920 \times 1080$ monocular webcam, transmitting images to a ground station equipped with an i7-13700KF CPU and RTX 4080 GPU via a 60Hz real-time onboard digital image transmission system. The real-world experimental results are presented in Figure \ref{fig:real}. The exploration trajectories are shown as the see. To demonstrate that the robot converged to the correct pose, the captured image at the final pose provided by camera and compared it with the corresponding target image. 

% Given images from a 1080p monocular webcam, our proposed BEINGS operates in real-time on the blimp. 

\section{Conclusion}
We introduced BEINGS, a novel approach for Image-goal navigation that addresses the limitations of existing methods by combining the strengths of learning-based and exploration-based strategies. Our experimental results demonstrate it outperforms existing methods, achieving improved navigation efficiency and adaptability with reduced computational and data requirements.  We demonstrate its feasibility on a real-world robotic platform. Future work will explore adaptive particle filtering for self-localization~\cite{meng2024nurf}, extend BEINGS to outdoor environments, and leverage GPU acceleration to enhance computational efficiency.
% \section*{Acknowledgement}

\bibliographystyle{IEEEtran}
\bibliography{root}

\end{document}